\documentclass{article}

 \usepackage[preprint]{neurips_2025}

\usepackage[utf8]{inputenc} 
\usepackage[T1]{fontenc}    
\usepackage{booktabs}       
\usepackage{amsfonts}       
\usepackage{nicefrac}       
\usepackage{microtype}      
\usepackage{xcolor}         
\usepackage{graphicx}
\usepackage{placeins}
\usepackage{amsmath}
\usepackage{amssymb}
\usepackage{algorithm}
\usepackage{float}
\usepackage[hidelinks]{hyperref}

\title{RASPRef: Retrieval-Augmented Self-Supervised Prompt Refinement for Large Reasoning Models}

\author{%
  Rahul Soni \\
  Independent Researcher \\
  \texttt{sonirahulsr1@gmail.com} \\
}

\begin{document}
\maketitle

\begin{abstract}
  Recent reasoning-focused language models such as DeepSeek R1 and OpenAI o1 have demonstrated strong performance on structured reasoning benchmarks including GSM8K, MATH, and multi-hop question answering tasks.  However, their performance remains highly sensitive to prompt formulation, and designing effective prompts is typically a manual and iterative process that does not scale well across tasks or domains. To address this limitation, we introduce Retrieval-Augmented Self-Supervised Prompt Refinement (RASPRef), a framework that improves prompts without requiring human annotations or task-specific supervision. The approach retrieves relevant examples and previously generated reasoning trajectories, and leverages signals such as multi-sample consistency, verifier feedback, and model-generated critiques to iteratively refine the prompt. Unlike prior approaches that focus primarily on improving model outputs, RASPRef directly treats the prompt as the optimization target and improves it through an iterative retrieval-guided refinement process. Experiments on GSM8K-style mathematical reasoning tasks show that retrieval-guided prompting improves performance compared with a static prompting baseline. We further discuss how retrieval quality, trajectory selection, and self-supervised feedback signals may influence the effectiveness of prompt refinement. These findings suggest that prompt design remains a critical factor for reasoning-oriented language models, and that self-improving prompts offer a practical and scalable strategy for improving reasoning performance. 
\end{abstract}

\section{Introduction}
Recent advances in large language models, particularly reasoning-focused models, have significantly improved performance on tasks that require structured multi-step reasoning. These models now perform well on tasks such as mathematical reasoning, code generation, and multi-hop question answering, reflecting broader progress reported in recent research on reasoning-oriented language models \citep{wei2022chain, patil2025advancingreasoning}. Despite these advances, model performance still depends heavily on the prompts used to guide their reasoning process. Even small changes in wording, structure, or example selection can lead to substantial differences in model behavior, a limitation widely documented in prior work on prompt sensitivity and optimization \citep{yang2024llmoptimizers, srivastava2025revisiting, wang2024promptagent}.
Recent developments in retrieval-augmented generation and self-refinement methods highlight the growing need for automatic prompt refinement. Retrieval-augmented approaches allow models to incorporate information from external sources such as documents, past interactions, and task-specific corpora, improving grounding and reducing hallucinations \citep{li2025retrievalaugmentedlearning, luo2025rallrec}. At the same time, self-refinement and self-evaluation methods demonstrate that models can critique, revise, and stabilize their own outputs through iterative feedback loops, leading to improved clarity and correctness without additional training \citep{xiang2025selfsupervised}. Modern LLM applications also produce extensive interaction logs including queries, prompts, reasoning traces, and responses which represent a valuable but underused resource for improving prompt design over time \citep{wu2024surveyllmrec}.
Building on these developments, we propose Retrieval-Augmented Self-Supervised Prompt Refinement (RASPRef), a framework that combines retrieval of prior reasoning trajectories with self-supervised evaluation signals to automatically refine prompts. Instead of treating prompts as fixed instructions, RASPRef views them as adaptive components that evolve as additional reasoning trajectories are collected. The framework leverages the model's accumulated reasoning history to guide the construction of improved prompts for new inputs, aligning with recent research on retrieval-guided reasoning and reflective optimization \citep{yang2024llmoptimizers, wang2024promptagent}.
In this work, we present the formulation, methodology, and empirical evaluation of the RASPRef framework. Our evaluation demonstrates that the framework provides a practical and scalable approach to prompt optimization while operating without human labels or task-specific annotation, addressing limitations identified in recent research on prompt engineering and reasoning-based decision systems \citep{lester2021prompt, dubey2024llama3}.

\section{Related Work}

\subsection{Prompt Engineering and Optimization}
Prompt engineering has been explored through a wide range of optimization strategies for improving language model performance. Early work explored soft prompt tuning and embedding-level adjustments \citep{li2021prefix, lester2021prompt, hu2022lora},  approaches that require access to model weights and internal representations. Other research has developed gradient-free optimization, including search-based methods \citep{shin2020autoprompt, wen2023promptsearch}, evolutionary strategies \citep{guo2024prompt, fernando2024evolution, yang2024llmoptimizers}, and reinforcement-learning-based refinement pipelines \citep{deng2022rlprompt, zhang2023promptrl}.

While effective, these approaches often depend on labeled data, curated prompt pools, or reward models, and they typically require substantial computational resources or human supervision. Moreover, many techniques are incompatible with closed-source LLMs accessed only through  APIs, where gradients and internal model states are unavailable \citep{zhou2025multiagentdesign, pryzant2023prompt}. These limitations motivate the development of black-box, iterative methods for prompt refinement that do not rely on internal model access.

\subsection{Retrieval-Augmented Generation}
Retrieval-Augmented Generation (RAG) enhances large language models  by grounding model outputs in retrieved external evidence \citep{lewis2020rag}. Early systems emphasized retrieval of short text passages \citep{izacard2021fid}, while later work expanded retrieval to tool demonstrations, program traces, and previous dialogue turns \citep{luo2025rallrec}. RAG has proven effective across knowledge-intensive tasks such as open-domain question answering \citep{karpukhin2020dense} and reasoning-centric dialogue systems \citep{shuster2021retrieval}.  

These developments show that retrieval supports richer reasoning, reduces hallucinations, and enables models to incorporate information that is not stored in their parameters. Recent work also integrates retrieval with recommendation systems \citep{wu2024surveyllmrec} and decision-making pipelines \citep{li2025retrievalaugmentedlearning}, underscoring the broader importance of retrieval-augmented inference.

\subsection{Self-Improvement and Test-Time Adaptation}
Self-improvement methods allow models to revise their own outputs through iterative feedback cycles during inference. Self-critique and reflection frameworks \citep{shinn2023reflexion, madaan2023selfrefine} refine responses through repeated generation–critique–revision loops, while consistency-based filtering may further improve robustness \citep{wang2023selfconsistency}. Verifier-based approaches further evaluate reasoning quality using auxiliary scoring models \citep{uesato2022processsupervision}. 

More recent work explores structured multi-agent critique systems  \citep{agarwal2024agents, yin2025multiagent}, memory-augmented reflection mechanisms \citep{dziri2024memory}, and self-play-style reasoning frameworks  \citep{openai2024reasoning}. However, most of these methods focus on improving generated answers rather than improving the prompts themselves. As a result, prompt quality remains a key bottleneck, with relatively little work dedicated to systematic, label-free prompt refinement. 

\subsection{Reasoning Models and Chain-of-Thought}
Reasoning-focused models such as DeepSeek-R1 \citep{guo2025deepseek} and OpenAI o1 \citep{zhong2024o1} rely heavily on detailed multi-step reasoning traces, often framed as chain-of-thought (CoT) reasoning \citep{wei2022chain}. CoT-based prompting improves performance on arithmetic, symbolic, and multi-hop reasoning tasks, yet these models remain highly sensitive to prompt phrasing, structure, and example selection \citep{suzgun2022cot, srivastava2025revisiting}. 

Because instruction clarity and example selection strongly influence reasoning behavior, large reasoning models (LRMs) are natural candidates for systematic prompt refinement. Prior work suggests that structured or iterative prompting such as MCTS-based prompt search \citep{wang2024promptagent} or mutation-based rewriting \citep{yang2024llmoptimizers}  can substantially improve structured reasoning performance. 

\paragraph{RASPRef in Context.}
RASPRef builds on these research directions by combining retrieval of relevant solution trajectories \citep{lewis2020rag, luo2025rallrec} with self-supervised evaluation signals inspired by consistency-based reasoning \citep{wang2023selfconsistency}, verifier feedback \citep{uesato2022processsupervision}, and model-driven critique mechanisms \citep{madaan2023selfrefine}. Through this combination, RASPRef provides a unified framework for automatic prompt refinement that adapts to new task instances and improves as additional reasoning trajectories are accumulated.

\begin{figure}[H]
\centering
\includegraphics[width=\linewidth]{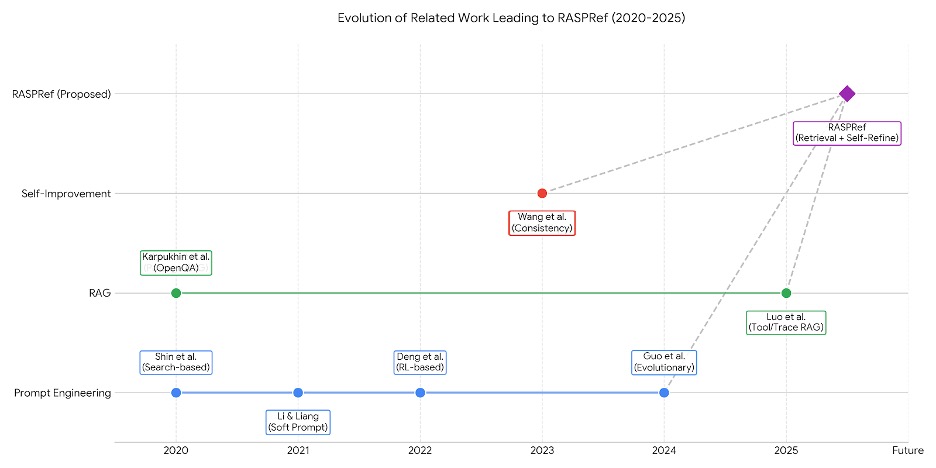}
\caption{Evolution of research directions leading to the RASPRef framework.}
\label{fig:raspref_framework}
\end{figure}

\section{Methodology}

\subsection{Problem Setup}

We consider a reasoning model with fixed parameters, similar to recent large reasoning systems used in complex decision-making and language tasks \citep{guo2025deepseek, zhong2024o1}. For each new input problem, the goal is to construct a prompt that encourages the model to produce clear, structured, and reliable reasoning. To support this objective, we maintain a collection of past reasoning trajectories, a setting that parallels prior work on retrieval-augmented reasoning and decision-making \citep{li2025retrievalaugmentedlearning, luo2025rallrec}.

Each stored trajectory contains the following elements:
\begin{itemize}
    \item a previous input instance,
    \item the prompt that was used for that input,
    \item the full reasoning trace generated by the model,
    \item an optional signal reflecting correctness or reward, and
    \item several self-supervised measures such as sample consistency or scores assigned by a verifier model \citep{shinn2023reflexion, madaan2023selfrefine, uesato2022processsupervision}.
\end{itemize}

As the system is used, this collection grows and records how the model behaves across a wide range of input problems. The objective is to use these accumulated trajectories to construct a refined prompt for a new input. We evaluate candidate prompts using a measure of prompt quality that does not require labeled data, consistent with recent work that treats evaluation signals as weak or self-supervised feedback \citep{seo2025promptoptimization, li2025retrievalaugmentedlearning}. This measure reflects the clarity, stability, and internal coherence of the model’s responses when guided by a particular prompt.

RASPRef refines prompts through a retrieval-guided and self-supervised optimization process. It is conceptually related to discrete prompt optimization and search-based methods that operate over natural language prompts \citep{yang2024llmoptimizers, seo2025promptoptimization, pryzant2023prompt}, but differs by explicitly combining retrieval over past trajectories with prompt-level self-refinement.

\paragraph{Step 1: Retrieval of Relevant Trajectories.}
For a new input, the system retrieves past trajectories whose input problems are most similar to the current task. This retrieval stage follows the general idea of retrieval augmented generation \citep{lewis2020rag}, where external memory is used to provide relevant context. Similarity may reflect task structure, domain characteristics, or the type of reasoning required. The retrieved trajectories form the foundation for constructing an initial prompt. 

\paragraph{Step 2: Construction of the Initial Prompt.}
Using the retrieved trajectories, RASPRef constructs an initial version of the prompt. This prompt contains:
\begin{itemize}
    \item general instructions that describe the task and specify the desired reasoning style,
    \item guidelines distilled from high-quality reasoning traces, and
    \item representative examples drawn from the retrieved trajectories.
\end{itemize}

The distilled guidelines are obtained by summarizing patterns that appear consistently in successful reasoning traces, a process related to the extraction of reusable reasoning assets or policies in previous systems \citep{seo2025promptoptimization, li2025retrievalaugmentedlearning}.

\paragraph{Step 3: Self-Supervised Evaluation.}
Given a candidate prompt, the model generates multiple responses for the same input. These responses are evaluated using self-supervised signals that do not rely on ground-truth labels. The signals include:
\begin{itemize}
    \item agreement across multiple samples, following ideas from self-consistency and test-time scaling \citep{wang2023selfconsistency, li2025retrievalaugmentedlearning},
    \item scores from a verifier model that judges plausibility and structure \citep{uesato2022processsupervision},
    \item self-critique, in which the model analyzes its own reasoning \citep{shinn2023reflexion, madaan2023selfrefine}, and
    \item how well the reasoning reuses relevant elements from the retrieved trajectories.
\end{itemize}

Together, these signals provide an estimate of prompt quality and identify aspects of the prompt that may be unclear or ineffective.

\paragraph{Step 4: Iterative Prompt Refinement.}
In the final stage, the model acts as an editor that revises and improves the prompt itself. Inspired by multi-agent and reflection-based optimization frameworks \citep{agarwal2024agents, yin2025multiagent}, the model reviews the prompt along with the generated responses and proposes targeted revisions that improve clarity, structure, and error avoidance. The revised prompt is then evaluated again with the same self-supervised signals, and the process is repeated. 

The refinement loop continues until the prompt stabilizes or a predefined computational budget is reached. In practice, most of the improvements are expected to occur within the first few refinement iterations, consistent with observations in prior work on search-based prompt optimization and retrieval-augmented self-improvement \citep{seo2025promptoptimization, li2025retrievalaugmentedlearning}.

\subsection{Prompt Quality Objective}

We formalize prompt refinement as an optimization problem over a discrete space of natural language prompts. Let $x \in \mathcal{X}$ denote an input problem, $y \in \mathcal{Y}$ a structured reasoning trace, and $p \in \mathcal{P}$ a prompt. A frozen reasoning model $f_{\theta}$ defines a distribution over reasoning traces:
\begin{equation}
y \sim f_{\theta}(\cdot \mid x, p).
\end{equation}

Since ground-truth labels are unavailable at test time, we introduce a self-supervised prompt quality score $Q(p; x)$ that aggregates multiple weak signals:
\begin{equation}
Q(p; x) =
\alpha \, C_{\mathrm{cons}}(p; x)
+ \beta \, C_{\mathrm{ver}}(p; x)
+ \gamma \, C_{\mathrm{crit}}(p; x)
+ \delta \, C_{\mathrm{ret}}(p; x),
\end{equation}
where:
\begin{itemize}
    \item $C_{\mathrm{cons}}$ measures multi-sample consistency across $K$ generated traces,
    \item $C_{\mathrm{ver}}$ is a verifier score that rates plausibility and structure,
    \item $C_{\mathrm{crit}}$ captures self-critique quality, and
    \item $C_{\mathrm{ret}}$ measures alignment with retrieved trajectories.
\end{itemize}

Concretely, for a given prompt $p$ and input $x$, we draw $K$ samples
\begin{equation}
\{y^{(k)}\}_{k=1}^{K} \sim f_{\theta}(\cdot \mid x, p),
\end{equation}
and define
\begin{equation}
C_{\mathrm{cons}}(p; x)
=
\frac{1}{K(K-1)}
\sum_{k \neq k'}
\mathbf{1}\!\left[\phi\!\left(y^{(k)}\right)=\phi\!\left(y^{(k')}\right)\right],
\end{equation}
where $\phi(\cdot)$ extracts a normalized final answer or canonicalized program.

The verifier score is
\begin{equation}
C_{\mathrm{ver}}(p; x)
=
\frac{1}{K}
\sum_{k=1}^{K}
v_{\psi}(x, y^{(k)}),
\end{equation}
where $v_{\psi}$ is a lightweight verifier model returning a scalar in $[0,1]$.

For self-critique, the reasoning model produces a critique $c^{(k)}$ for each trace $y^{(k)}$; a scoring function $s_{\mathrm{crit}}$ converts these into
\begin{equation}
C_{\mathrm{crit}}(p; x)
=
\frac{1}{K}
\sum_{k=1}^{K}
s_{\mathrm{crit}}(x, y^{(k)}, c^{(k)}).
\end{equation}

Finally, $C_{\mathrm{ret}}$ measures how effectively the trace reuses and cites retrieved reasoning assets $T_x$:
\begin{equation}
C_{\mathrm{ret}}(p; x)
=
\frac{1}{K}
\sum_{k=1}^{K}
s_{\mathrm{ret}}(y^{(k)}, T_x),
\end{equation}
where $s_{\mathrm{ret}}$ rewards explicit references to relevant steps or subgoals found in $T_x$.

The overall goal of RASPRef is to iteratively construct a refined prompt $p^\star$ that approximately maximizes the expected quality under the test-time input distribution:
\begin{equation}
p^\star
\approx
\arg\max_{p \in \mathcal{P}}
\; \mathbb{E}_{x \sim \mathcal{D}_{\mathrm{test}}}
\left[ Q(p; x) \right].
\end{equation}

This optimization uses only retrieval over past trajectories and self-supervised signals.

\subsection{RASPRef Algorithm}
Algorithm~\ref{alg:raspref} summarizes the refinement loop. The procedure treats the reasoning model as a black box and operates purely on text prompts and generated trajectories.

\begin{algorithm} [t]
\caption{RASPRef: Retrieval-Augmented Self-Supervised Prompt Refinement}
\label{alg:raspref}
\begin{small}
\textbf{Input:} reasoning model $f_{\theta}$, verifier $v_{\psi}$, trajectory store $\mathcal{M}$, base prompt $p_0$, retrieval function $\textsc{Retrieve}(\cdot)$, maximum refinement rounds $R$, samples per round $K$ \\
\textbf{Output:} refined prompt $p_R$
\begin{enumerate}
    \item For each new input problem $x$:
    \begin{enumerate}
        \item Retrieve relevant trajectories:
        \[
        T_x \leftarrow \textsc{Retrieve}(x, \mathcal{M})
        \]
        \item Construct initial prompt:
        \[
        p^{(0)} \leftarrow \textsc{BuildPrompt}(p_0, T_x)
        \]
        \item For $r = 0, \dots, R-1$:
        \begin{enumerate}
            \item Generate reasoning traces:
            \[
            \{y_r^{(k)}\}_{k=1}^{K} \sim f_{\theta}(\cdot \mid x, p^{(r)})
            \]
            \item Compute self-supervised score:
            \[
            q_r \leftarrow Q(p^{(r)}; x)
            \]
            \item Generate critiques and prompt edits:
            \[
            e_r \leftarrow \textsc{CritiqueAndEdit}(p^{(r)}, x, \{y_r^{(k)}\}, T_x)
            \]
            \item Apply edits:
            \[
            p^{(r+1)} \leftarrow \textsc{ApplyEdits}(p^{(r)}, e_r)
            \]
            \item If $Q(p^{(r+1)}; x) \leq q_r$, stop early.
        \end{enumerate}
        \item Output refined prompt:
        \[
        p_R \leftarrow p^{(r)}
        \]
        \item Store the new trajectory and feedback in $\mathcal{M}$.
    \end{enumerate}
\end{enumerate}
\end{small}
\end{algorithm}

The helper routines \textsc{BuildPrompt}, \textsc{CritiqueAndEdit}, and \textsc{ApplyEdits} are implemented using the same underlying language model. Prompts are edited through natural-language instructions such as ``clarify step-by-step reasoning structure,'' ``discourage premature final answers,'' or ``emphasize verification against retrieved examples.''

Figure~2 illustrates the overall RASPRef workflow, including retrieval of prior trajectories, prompt construction, reasoning generation, scoring, and iterative refinement.

\begin{figure}[t]
\centering
\includegraphics[width=0.82\columnwidth]{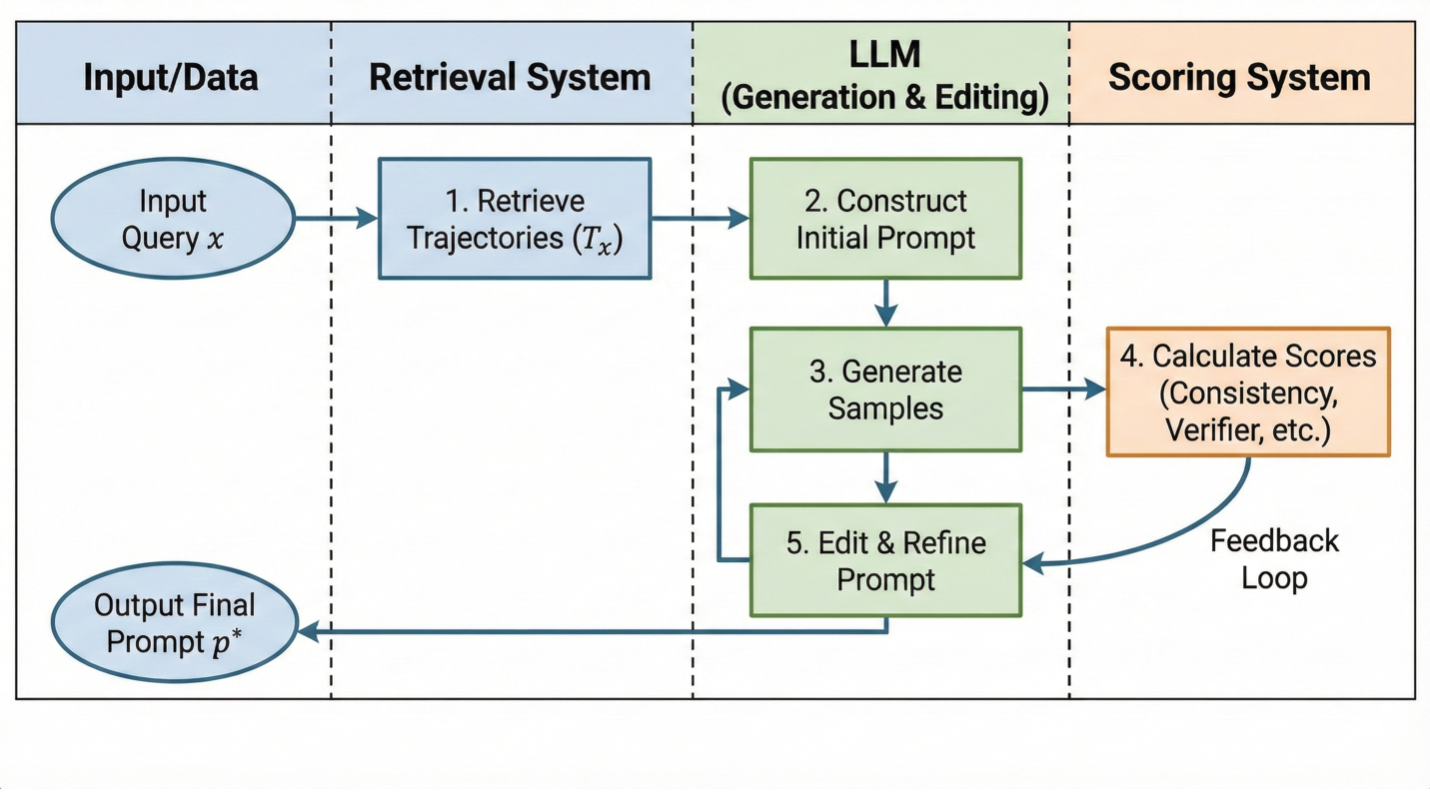}
\caption{Iterative prompt refinement workflow of the RASPRef framework.}
\label{fig:raspref_workflow}
\end{figure}

\FloatBarrier
\section{Experiments}

This section evaluates the RASPRef prototype on mathematical reasoning tasks.
The goal of the evaluation is to examine whether retrieval-guided prompt
construction improves answer accuracy compared with a static prompting
baseline. The experiments focus on arithmetic reasoning using the GSM8K
dataset.

Because the current prototype primarily evaluates the retrieval-guided prompt
construction component, the empirical results should be interpreted as an
initial validation of the broader RASPRef framework rather than a full
end-to-end evaluation of iterative self-supervised prompt refinement.

\subsection{Experimental Setup}

\paragraph{Model.}
The prototype uses \textbf{gpt-4o-mini}, accessed through an API, and treats
the model as a frozen reasoning model during inference. Prompting follows a
chain-of-thought style configuration for multi-step reasoning. Unless otherwise
specified, decoding uses temperature $=0.7$ and top-$p=0.95$.

\paragraph{Hardware and Runtime Environment.}
All experiments are conducted on a local development machine using a CPU-based
Python environment. Retrieval, prompt construction, and evaluation are
implemented in Python. Because the reasoning model is accessed through an API,
no local GPU training is required.

\paragraph{Retrieval Infrastructure.}
Prior problem-solution trajectories are encoded using
\textbf{text-embedding-3-small} and stored in a vector index. For each new
problem instance, the system retrieves the top-$k$ most relevant prior
trajectories using cosine similarity, where $k=5$, and
incorporates them into the prompt as contextual examples.

\paragraph{Dataset Preparation.}
Experiments are conducted using arithmetic reasoning problems derived from the
\textbf{GSM8K} benchmark. GSM8K contains grade-school math word problems that
require multi-step reasoning and numerical computation. For the evaluation, we
use a subset of 500 test examples. In the retrieval-based setting, the
retrieval index is constructed from 200 training examples.

\paragraph{Evaluation Protocol.}
For each test instance, the model generates a chain-of-thought reasoning trace
and a final answer. The final answer is extracted from the generated output and
compared against the reference solution to determine correctness. We report
accuracy over the evaluated subset.

\subsection{Baselines}

To analyze the effect of retrieval, we compare two prompting configurations:

\begin{itemize}
\item \textbf{Static Prompting}: a fixed chain-of-thought prompt template used
for all examples without retrieval.
\item \textbf{Retrieval-Augmented Prompting}: prompts augmented with retrieved
problem--solution trajectories selected using embedding similarity.
\end{itemize}

\subsection{Results}

Both prompting settings are evaluated on a 500-example subset of GSM8K. The
results are summarized in Table~\ref{tab:gsm8k_results}.

\begin{table}[H]
\centering
\caption{Evaluation results on a 500-example GSM8K subset.}
\label{tab:gsm8k_results}
\begin{tabular}{lcc}
\toprule
\textbf{Setting} & \textbf{Samples} & \textbf{Accuracy} \\
\midrule
Static prompting & 500 & 85.6\% \\
Retrieval-augmented prompting & 500 & 95.0\% \\
\bottomrule
\end{tabular}
\end{table}

Retrieval-augmented prompting improves accuracy compared with the static
prompting baseline. The results indicate that retrieved reasoning trajectories
provide useful contextual guidance that helps the model structure intermediate
reasoning steps more effectively.

\subsection{Discussion}

The experimental results demonstrate that retrieval-guided prompting can
improve performance on multi-step arithmetic reasoning tasks. Retrieved
trajectories appear to assist the model in organizing intermediate reasoning
steps and reducing reasoning errors.

The current evaluation focuses on a prototype implementation and a single
reasoning benchmark. Future work will expand the evaluation to larger datasets
and additional reasoning domains.

\FloatBarrier
\section{Further Analysis}

This section provides additional insights into the behavior, limitations, and practical considerations of the RASPRef framework beyond the main experimental evaluation.

\subsection{Influence of Trajectory Quality}

The quality of retrieved trajectories can have a direct impact on the stability and effectiveness of the RASPRef framework. Prior work has shown that noisy reasoning traces can degrade model behavior during self-refinement \citep{madaan2023selfrefine, shinn2023reflexion}. 

Trajectories that contain clear intermediate reasoning steps are more likely to produce useful prompt refinements, whereas trajectories containing errors or inconsistent logic may bias the prompt in undesirable ways. Filtering trajectories using verifier-based scoring, similar to prior work on process supervision, may therefore improve the stability of the refinement process \citep{uesato2022processsupervision}. Consistency-based filtering across multiple sampled solutions may provide an additional signal for reliability \citep{wang2023selfconsistency}. 

These observations highlight the importance of maintaining a well-curated trajectory store for reliable prompt optimization.

\subsection{Generalization to Rare and Long-Tail Cases}

The design of RASPRef is intended to support generalization on rare or structurally unusual inputs. Retrieval mechanisms can identify past trajectories that share deeper reasoning patterns rather than superficial lexical similarity, consistent with observations reported in retrieval-augmented reasoning systems \citep{karpukhin2020dense}.

This mechanism may enable the framework to transfer high-level reasoning structures even when the test input falls outside the typical distribution of earlier tasks. Compared with static prompts or nearest-example prompting, retrieval-guided prompt construction may therefore help reasoning models adapt to long-tail cases where stable reasoning chains are difficult to form.

\subsection{Risks of Overfitting}

Although iterative refinement has the potential to improve prompt quality, it also introduces the risk of narrow overfitting when too many samples from the same neighborhood are used during prompt refinement. Similar risks have been discussed in iterative prompt search and self-improvement frameworks \citep{kojima2022zeroshot, yang2024selfimprove}.

Excessive refinement may cause prompts to encode overly specific instructions that fail to generalize to new tasks. Limiting the number of refinement rounds and encouraging concise prompt edits may help mitigate this form of overspecialization.

\subsection{Scalability and Efficiency}

RASPRef introduces computational overhead because each refinement cycle involves additional sampling and scoring. However, the use of smaller verifier models can help mitigate this cost.

This design follows trends in efficient evaluation pipelines where lightweight models provide supervision signals at reduced computational expense \citep{uesato2022processsupervision}. Such an approach may enable RASPRef-style refinement loops to remain practical even when applied to larger reasoning models.

\subsection{Interpretability of Refined Prompts}

A key advantage of RASPRef is that the refined prompts remain transparent and interpretable. Unlike embedding-level optimization or reinforcement learning approaches that modify model parameters implicitly \citep{li2021prefix, hu2022lora}, RASPRef produces explicit prompt revisions that reveal the heuristics extracted from prior reasoning traces.

The refined prompts can incorporate clearer task descriptions, more structured reasoning guidance, and distilled patterns derived from earlier trajectories. This transparency provides insight into how reasoning models organize multi-step problem solving, consistent with findings from research on interpretable chain-of-thought prompting \citep{wei2022chain}.

\section{Conclusion}

We introduced \textbf{RASPRef}, a retrieval-augmented framework for prompt
refinement in reasoning-oriented language models. By leveraging past reasoning
trajectories and lightweight self-evaluation signals, RASPRef treats prompts as
adaptive components that can improve as additional reasoning experience is
accumulated. The framework combines retrieval of past reasoning trajectories
with self-supervised evaluation signals to iteratively improve prompts during
inference.

In our experiments on GSM8K-style mathematical reasoning tasks, retrieval-guided
prompt construction substantially outperformed a static prompting baseline. On a
500-problem evaluation subset, accuracy improved from 85.6\% with static
prompting to 95.0\% with retrieval-augmented prompting. These results suggest
that incorporating previously successful reasoning trajectories provides useful
contextual guidance that helps models structure multi-step reasoning more
effectively.

Overall, the findings indicate that retrieval-based prompt refinement offers a
promising direction for improving reasoning performance without requiring model
fine-tuning or additional human annotations. Future work may extend the
framework to broader reasoning domains, integrate stronger verification
signals, and evaluate retrieval-guided refinement on additional benchmarks
such as multi-hop question answering and program synthesis.

\bibliographystyle{abbrvnat}
\bibliography{references}

\end{document}